\documentclass[sigconf,natbib=true]{acmart}

\AtBeginDocument{%
  }

\setcopyright{acmlicensed}
\copyrightyear{2026}
\acmYear{2026}
\setcopyright{cc}
\setcctype{by}
\acmConference[SIGIR '26]{Proceedings of the 49th International ACM SIGIR Conference on Research and Development in Information Retrieval}{July 20--24, 2026}{Melbourne, VIC, Australia}
\acmBooktitle{Proceedings of the 49th International ACM SIGIR Conference on Research and Development in Information Retrieval (SIGIR '26), July 20--24, 2026, Melbourne, VIC, Australia}
\acmDOI{10.1145/3805712.3808602}
\acmISBN{979-8-4007-2599-9/2026/07}

\usepackage{enumitem}
\usepackage{xspace}
\usepackage{hyperref}
\usepackage{multirow}
\usepackage{pifont}
\usepackage{makecell}
\newcommand{\system}{PeerPrism}

\usepackage[most]{tcolorbox}

\begin{document}

\title{\system: Peer Evaluation Expertise vs Review-writing AI }


 \author{Soroush Sadeghian}
 \orcid{0009-0005-2172-7617}
 \affiliation{%
   \institution{Reviewerly}
 \city{Toronto}
 \state{Ontario}
 \country{Canada}
   }
 \email{soroushsa@reviewer.ly}

 \author{Alireza Daghighfarsoodeh}
 \orcid{0009-0007-8806-121X}
 \affiliation{%
   \institution{Reviewerly}
  \city{Toronto}
  \state{Ontario}
  \country{Canada}
   }
 \email{alirezada@reviewer.ly}
 
 \author{Radin Cheraghi}
 \orcid{0009-0006-0887-2614}
 \affiliation{%
   \institution{Reviewerly}
 \city{Toronto}
 \state{Ontario}
 \country{Canada}
   }
 \email{radinch@reviewer.ly}

 \author{Sajad Ebrahimi}
 \authornote{Corresponding author}
 \orcid{0009-0003-1630-3938}
 \affiliation{%
   \institution{University of Toronto, Reviewerly}
 \city{Toronto}
 \state{Ontario}
  \country{Canada}
   }
 \email{s.ebrahimi@utoronto.ca}

\author{Negar Arabzadeh}
 \orcid{0000-0002-4411-7089}
 \affiliation{%
   \institution{UC Berkeley, Reviewerly}
  \city{Berkeley}
  \state{California}
  \country{United States}
   }
 \email{negara@berkeley.edu}

 \author{Ebrahim Bagheri}
 \orcid{0000-0002-5148-6237}
 \affiliation{%
   \institution{University of Toronto, Reviewerly}
 \city{Toronto}
 \state{Ontario}
 \country{Canada}
 }
 \email{ebrahim.bagheri@utoronto.ca}

\renewcommand{\shortauthors}{Soroush Sadeghian et al.}

\begin{abstract}
  Large Language Models (LLMs) are increasingly used in scientific peer review, assisting with drafting, rewriting, expansion, and refinement. However, existing peer-review LLM detection methods largely treat authorship as a binary problem-human vs.\ AI-without accounting for the hybrid nature of modern review workflows. In practice, evaluative ideas and surface realization may originate from different sources, creating a spectrum of human-AI collaboration.

In this work, we introduce \system, a large-scale benchmark of 20,690 peer reviews explicitly designed to disentangle idea provenance from text provenance. We construct controlled generation regimes spanning fully human, fully synthetic, and multiple hybrid transformations. This design enables systematic evaluation of whether detectors identify the origin of the surface text or the origin of the evaluative reasoning.
We benchmark state-of-the-art LLM text detection methods on \system. While several methods achieve high accuracy on the standard binary task (human vs.\ fully synthetic), their predictions diverge sharply under hybrid regimes. In particular, when ideas originate from humans but the surface text is AI-generated, detectors frequently disagree and produce contradictory classifications. Accompanied by stylometric and semantic analyses, our results show that current detection methods conflate surface realization with intellectual contribution.

Overall, we demonstrate that LLM detection in peer review cannot be reduced to a binary attribution problem. Instead, authorship must be modeled as a multidimensional construct spanning semantic reasoning and stylistic realization. \system\ is the first benchmark evaluating human-AI collaboration in these settings. We release all code, data, prompts, and evaluation scripts to facilitate reproducible research at \url{https://github.com/Reviewerly-Inc/PeerPrism}.
\end{abstract}

\ccsdesc[500]{Computing methodologies~Natural language processing}

\begin{abstract}
    Large Language Models (LLMs) are increasingly used in scientific peer review, assisting with drafting, rewriting, expansion, and refinement. However, existing peer-review LLM detection methods largely treat authorship as a binary problem-human vs.\ AI-without accounting for the hybrid nature of modern review workflows. In practice, evaluative ideas and surface realization may originate from different sources, creating a spectrum of human-AI collaboration.

In this work, we introduce \system, a large-scale benchmark of 20,690 peer reviews explicitly designed to disentangle idea provenance from text provenance. We construct controlled generation regimes spanning fully human, fully synthetic, and multiple hybrid transformations. This design enables systematic evaluation of whether detectors identify the origin of the surface text or the origin of the evaluative reasoning.
We benchmark state-of-the-art LLM text detection methods on \system. While several methods achieve high accuracy on the standard binary task (human vs.\ fully synthetic), their predictions diverge sharply under hybrid regimes. In particular, when ideas originate from humans but the surface text is AI-generated, detectors frequently disagree and produce contradictory classifications. Accompanied by stylometric and semantic analyses, our results show that current detection methods conflate surface realization with intellectual contribution.

Overall, we demonstrate that LLM detection in peer review cannot be reduced to a binary attribution problem. Instead, authorship must be modeled as a multidimensional construct spanning semantic reasoning and stylistic realization. \system\ is the first benchmark evaluating human-AI collaboration in these settings. We release all code, data, prompts, and evaluation scripts to facilitate reproducible research at \url{https://github.com/Reviewerly-Inc/PeerPrism}.
\end{abstract}

\keywords{Peer Review, AI-generated Text Detection, Human-AI Collaboration}


\maketitle

\vspace{-0.5em}
\section{Introduction}

Peer review is a cornerstone of scientific progress, serving as a mechanism for quality assurance, expert feedback, and community calibration \cite{hillard2021peer, kelly2014peer, arabzadeh2024reviewerly, ebrahimi2025exharmony}. At the same time, Large Language Models (LLMs) have become embedded across scholarly workflows, from drafting and editing to summarization and idea exploration \cite{ahn2024transformative, van2023chatgpt}. Their fluency makes them attractive tools for assisting review writing \cite{lee2025role, arabzadeh2026can}. However, as LLMs increasingly participate not only in writing science but in judging it, the stakes fundamentally change. Peer review determines which ideas are published, funded, and amplified. If reviews are shallow, biased, hallucinated, or misaligned with disciplinary standards, scientific decisions may be distorted \cite{zhu2025evaluating}. Integrating LLMs into peer review is therefore not merely a tooling issue, but a question of preserving the integrity and calibration of the scientific enterprise \cite{rao2025detecting, arabzadeh2025building, ebrahimi2025rottenreviews}.

More broadly, AI-generated text detection has become prevalent across domains, including education, journalism, hiring, and online content moderation \cite{sadasivan2023can, weber2023testing}. These systems aim to distinguish human-authored from machine-generated text. In peer review, similar detectors have been proposed to identify LLM-generated reviews \cite{liang2024can, rao2025detecting}. However, existing evidence suggests that LLMs exhibit distinctive stylistic properties-such as high fluency, structured organization, and characteristic lexical patterns-that may differ from typical human writing \cite{reinhart2025llms}. This raises a fundamental question: are current detectors capturing the substantive reasoning and evaluative quality of a review, or merely identifying surface-level stylistic signals?
This question is particularly consequential in peer review, where reviews are semi-structured, domain-specific, and grounded in expert judgment. They reference disciplinary norms (e.g., novelty, empirical rigor, ablations, baselines) and are tightly coupled to a specific target paper. Moreover, many venues permit LLM use for drafting or language refinement. In such cases, the underlying evaluative ideas may remain human, even if the written expression is AI-assisted. If detectors rely primarily on stylistic cues rather than domain-aware reasoning signals, human-authored reviews revised with LLM support may be incorrectly flagged. Ensuring that detection methods disentangle intellectual contribution from linguistic form and remain robust under domain shift is therefore essential in this high-stakes setting.

Crucially, review creation lies on a spectrum rather than a binary divide. The evaluative ideas may originate from either a human or an LLM, and the written expression may likewise be authored or revised by either. This yields multiple configurations beyond the simple ``human vs. AI'' framing commonly assumed in current benchmarks \cite{liang2024monitoring, lee2025role, hutson2025human}. While existing detection datasets typically focus on distinguishing fully synthetic from fully human reviews \cite{queralt2025ai}, they do not examine whether systems can identify the origin of the underlying evaluative reasoning. A robust evaluation framework should therefore account for this idea-text spectrum, rather than assuming that surface-level textual style faithfully reflects intellectual authorship.

In this work, we ask a fundamental question: \textit{do current LLM-detection tools distinguish the origin of evaluative ideas, or do they primarily rely on surface-level textual signals? }In other words, are detectors sensitive to who generated the reasoning behind a review, or merely to stylistic artifacts associated with LLM writing? Furthermore, if these two dimensions differ, how do existing detectors behave in the “gray area” where evaluative ideas originate from a human reviewer but the text is refined or expanded using an LLM?
This gray area itself spans a spectrum. LLM assistance may range from minor language polishing to structural reorganization, expansion, or partial drafting. In many realistic scenarios, the intellectual content remains human while the expression is AI-assisted. Understanding detector behavior across this continuum is essential for fair and reliable deployment.

To address these questions, we introduce \textbf{\system}, a dataset and benchmark designed to evaluate LLM-detection tools on scientific peer reviews under controlled generation regimes. We curate human-authored reviews from OpenReview-hosted venues (ICLR and NeurIPS) and construct multiple LLM-assisted variants that explicitly disentangle \emph{idea origin} from \emph{text origin}. This controlled design enables systematic analysis of detector behavior across realistic drafting, editing, and hybrid authorship scenarios, moving beyond the standard binary “human vs. AI” framing.

We evaluate representative detector baselines spanning likelihood based, perturbation based, embedding based, and supervised paradigms, including DetectGPT~\cite{mitchell2023detectgpt}, Fast-DetectGPT~\cite{bao2024fastdetectgpt}, Lastde++~\cite{xu2025trainingfree}, and RADAR~\cite{hu2023radarrobustaitextdetection}, across all \system\ review types. While several detectors achieve high accuracy on purely human and purely LLM-generated reviews, their predictions become unstable-and often contradictory-when human-originated ideas are expressed through LLM-generated text. These results expose fundamental limitations of current binary attribution methods in realistic, hybrid peer-review settings.
In summary, we make the following contributions:
\begin{itemize}[leftmargin=*, noitemsep, topsep=0pt]
\item \textbf{Task formulation:} We introduce the task of \emph{idea-text provenance disentanglement} in peer review, moving beyond the standard binary “human vs. AI” detection paradigm.

\item \textbf{Dataset and benchmark:} We release \system, a JSON-based dataset of peer reviews with controlled generation regimes that systematically separate idea provenance from text provenance, enabling fine-grained evaluation under realistic hybrid scenarios.

\item \textbf{Comprehensive detector benchmarking:} We evaluate representative likelihood-based, perturbation-based, embedding-based, and supervised LLM-detection baselines on this task, providing the first systematic study of detector behavior under mixed authorship conditions.

\item \textbf{Stylistic and semantic analysis:} We conduct stylometric and semantic analyses-including rhetorical structure, citation behavior, and embedding similarity-to characterize how human, fully synthetic, and transformed reviews differ, and to diagnose detector failure modes.

\end{itemize}
\vspace{-0.5em}
\section{Related Work}

\paragraph{LLM usage and detection in peer review}
Empirical studies confirm the widespread adoption of LLM assistance in peer review and the increasing difficulty of disentangling hybrid authorship~\cite{liang2024monitoring,shen2026detecting}. This prevalence has motivated domain-specific detection efforts. Yu et al.~\cite{Yu2025IsYP} introduce Anchor, a context-aware method that uses embedding similarity but assumes access to reference generations tied to the target manuscript. Closest to our setting, MixRevDetect~\cite{kumar2025mixrevdetect} presents a supervised framework for identifying AI-generated sentences within hybrid reviews. However, it relies on labeled training data and implicitly conflates AI-generated text with AI-generated ideas. The growing reliance on automated detection also raises ethical and governance concerns. Specifically, detection tools may disproportionately flag standard academic writing styles and encode systematic biases related to institutional prestige~\cite{elazar2026arxiv,vasu2025justice}.

General methods for detecting LLM-generated text and their associated benchmarks are discussed in Section~\ref{sec:benchmark}. In contrast to prior work, \system\ evaluates detectors within realistic workflows, explicitly disentangling the provenance of evaluative ideas from surface text. This separation is critical for accurately assessing detection reliability in hybrid authorship scenarios that increasingly reflect real-world practice.

\vspace{-0.5em}
\section{Dataset}
\label{sec:dataset}

To rigorously investigate AI attribution in scientific peer review, we introduce \system, a new corpus explicitly designed to disentangle \emph{idea origin} from \emph{text origin}. Unlike prior datasets that frame detection as a binary classification problem (Human vs.\ AI), \system\ models peer review as a hybrid environment in which evaluative ideas and their surface realization may stem from different sources. The dataset comprises 20,690 reviews derived from 160 seed papers and spans fully human, fully AI-generated, and systematically constructed hybrid provenance settings.

\subsection{ Data Collection}

We collect ground-truth human reviews from OpenReview, focusing on two top-tier machine learning venues: \emph{ICLR} and \emph{NeurIPS}~\cite{rao2025detecting}. To ensure temporal diversity and mitigate confounding effects related to model release timelines, we sample papers across four years (2021-2024). Notably, reviews from 2021 and 2022 predate the widespread availability of ChatGPT and similar LLMs~\cite{liang2024monitoring}, whereas those from 2023 and 2024 may reflect varying degrees of LLM-assisted writing. This temporal stratification enables us to capture both pre-LLM and post-LLM reviewing behaviors.

The dataset includes 160 papers in total, corresponding to 20 papers per venue per year. To avoid selection bias, paper sampling is balanced by decision outcome, with 10 accepted and 10 rejected papers in each venue-year subset. For every selected paper, we collect the full manuscript (PDF content), associated metadata, and all available human-written reviews.
To standardize document representation, we convert each manuscript into structured Markdown using Microsoft MarkItDown~\cite{simmering2025meet}\footnote{\url{https://github.com/microsoft/markitdown}}
. This conversion preserves document hierarchy and semantic structure while minimizing noise introduced by raw PDF extraction. The resulting structured format ensures consistent model inputs and improves the reliability of downstream idea extraction and review generation pipelines.
Overall, this process yields {674 human-written reviews}. These reviews serve both as the gold standard for human-origin ideas and human-authored text, and as seed material for the controlled transformation pipelines used to generate hybrid and AI-origin variants.

\subsection{Review Generation Regimes}
We use six frontier LLMs from proprietary and open-source families including, \texttt{OpenAI-GPT-5}, \texttt{OpenAI-o4-mini}, \texttt{Gemini-2.5-Flash}, \texttt{DeepSeek-R1}, \texttt{Llama-4-Scout}, and \texttt{Claude-Haiku-4.5}. We intentionally focus on high-capacity models rather than smaller baselines, as peer review requires long-context reasoning, domain-specific judgment, and structured critique. Smaller models often lack sufficient context length and evaluation quality to produce realistic reviews. Our objective is to evaluate detector behavior under frontier-level generation, where synthetic reviews are both plausible and difficult to detect.
In the following, we describe the review generation regimes implemented in \system.

\subsubsection{Fully Synthetic Reviews (AI Ideas / AI Text)} \label{sec:fullgen}
In this setting, the model functions as an independent reviewer. It reads the manuscript and generates a review \textit{de novo}, without access to any human-written reviews. To capture variability in reviewing style, inspired by \cite{tan2024phantom,kim2025persona,wu2023large}, we employ four reviewer personas across all generation settings: \textit{Conservative}, \textit{Highly-Detailed}, \textit{Lazy Reviewer}, and \textit{Nitpicky}. Each persona encodes a distinct rhetorical tone, level of analytical depth, and strictness, ensuring stylistic diversity independent of idea provenance. Detailed prompt specifications for each persona are available in our GitHub repository.

Crucially, reviews are generated per paper, not per human review. For each of the 160 papers, every model generates four reviews (one per persona). This results in 3,840 fully synthetic reviews ($160 \text{ papers} \times 6 \text{ models} \times 4 \text{ styles}$). These instances represent the baseline for AI-generated content where both the critical analysis (ideas) and the textual formulation originate from the model.

\subsubsection{Provenance-Controlled Reviews}
\label{sec:mixed}
To model hybrid authorship, we applied four transformation strategies to the 674 human source reviews. Each regime is applied across all six models, yielding 4,044 reviews per regime ($674 \times 6$). These strategies allow us to control the ``Idea Origin'' while keeping the ``Text Origin'' as AI.

\paragraph{1. Review Rewritten (Human Idea / AI Text)}
The model receives \textit{only} the human review text and is instructed to rewrite it stylistically without adding new information or accessing the manuscript. This isolates surface realization while preserving human inference.

\paragraph{2. Idea Extraction \& Review Regeneration (Human Idea / AI Text)}

\begin{table}[t]
\centering

\caption{Breakdown of the sub-datasets in \system\ by provenance and generation method.}

\label{tab:dataset_stats}
\scalebox{0.85}{
\begin{tabular}{lllr}
\toprule
\textbf{Data Partition} & \textbf{Idea Origin} & \textbf{Text Origin} & \textbf{Count} \\
\midrule
Human & Human & Human & 674 \\
Fully Synthetic & AI & AI & 3,840 \\
Rewritten & Human & AI & 4,044 \\
Expanded & Mixed & AI & 4,044 \\
Extract Regenerate & Human & AI & 4,044 \\
Hybrid & Mixed & AI & 4,044 \\
\midrule
\textbf{Total} & & & \textbf{20,690} \\
\bottomrule
\end{tabular}}
\end{table}

To explicitly disentangle evaluative reasoning from surface realization, we implement a strict two-stage pipeline~\cite{yuan2022can}. In the first stage, the model extracts the core evaluative ideas from the human review and manuscript, producing a structured representation~\cite{wang2020reviewrobot} of essential strengths, weaknesses, and concerns. This step isolates the semantic content--the ``gist'' of the review--while discarding stylistic features of the original human text.

In the second stage, a fresh model context generates a review using only this structured idea representation, without access to the original human wording or manuscript text. By regenerating the review from abstracted ideas rather than rewriting the human text, we eliminate stylistic anchoring to human expression~\cite{krishna2020reformulating}. As a result, the semantic core remains human-originated, while the surface realization is entirely AI-generated.
\paragraph{3. Review Expansion (Mixed Idea / AI Text)}

In this regime, the model receives both the original human review and the manuscript and is instructed to expand upon the reviewer’s critique~\cite{liang2024can}. The model elaborates on existing points, adds clarifications, and introduces additional supporting arguments grounded in the paper. Importantly, the expansion operates on a \emph{single} human review, preserving its original evaluative perspective~\cite{ghosal2019deepsentipeer} while extending it with additional detail. As a result, the core critique remains human-originated, but supplementary reasoning may be model-generated, yielding mixed idea provenance. The surface realization is fully AI-generated.
\paragraph{4. Hybrid Augmentation (Mixed Idea / AI Text)}

\begin{table*}[ht]
\centering
\small

\caption{Binary LLM detection performance (Human vs. Fully Synthetic). For each LLM, we report Accuracy and Macro-F1.}

\setlength{\tabcolsep}{7pt}
\renewcommand{\arraystretch}{1.15}

\begin{tabular}{l cc cc cc cc cc cc cc}
\hline
\textbf{method}
& \multicolumn{2}{c}{\textbf{Overall}}
& \multicolumn{2}{c}{\textbf{Gemini-2.5}}
& \multicolumn{2}{c}{\textbf{GPT-5}}
& \multicolumn{2}{c}{\textbf{o4-mini}}
& \multicolumn{2}{c}{\textbf{Claude-4.5}}
& \multicolumn{2}{c}{\textbf{DeepSeek-R1}}
& \multicolumn{2}{c}{\textbf{Llama-4}} \\
\cline{2-15}

& \textbf{Acc.} & \textbf{F1}
& \textbf{Acc.} & \textbf{F1}
& \textbf{Acc.} & \textbf{F1}
& \textbf{Acc.} & \textbf{F1}
& \textbf{Acc.} & \textbf{F1}
& \textbf{Acc.} & \textbf{F1}
& \textbf{Acc.} & \textbf{F1} \\
\hline

Anchor         
& 90.39 & 84.75
& 94.06 & 94.03 
& 95.51 & 95.49 
& 95.43 & 95.42 
& 96.65 & 96.64 
& 95.43 & 95.42 
& 87.60 & 87.34 \\

Binoculars     
& 30.99 & 30.96
& 56.90 & 47.66 
& 49.46 & 33.09 
& 49.46 & 33.09 
& 50.31 & 34.92 
& 51.07 & 36.54 
& 97.16 & 97.16 \\

DetectGPT      
& 53.25 & 46.97
& 35.77 & 29.91 
& 81.07 & 80.53 
& 76.86 & 76.52 
& 42.24 & 39.17 
& 74.43 & 74.18 
& 34.25 & 27.46 \\

Fast-DetectGPT 
& 89.45 & 82.72
& 94.29 & 94.29 
& 79.15 & 78.75 
& 89.57 & 89.56 
& 94.98 & 94.97 
& 88.13 & 88.10 
& 93.29 & 93.29 \\

GLTR           
& 30.59 & 27.55
& 37.82 & 37.76 
& 17.35 & 14.79 
& 22.68 & 21.54 
& 24.81 & 24.07 
& 24.51 & 23.71 
& 64.76 & 61.28 \\

Lastde++       
& 52.25 & 48.99
& 86.99 & 86.95 
& 51.37 & 41.44 
& 48.86 & 36.80 
& 90.18 & 90.18 
& 46.88 & 32.89 
& 86.65 & 86.60 \\

RADAR          
& 25.47 & 25.40
& 39.19 & 29.04 
& 38.51 & 27.80 
& 38.51 & 27.80 
& 38.66 & 28.08 
& 39.04 & 28.77 
& 86.27 & 86.16 \\

\hline
\end{tabular}

\label{tab:accuracy_by_model}
\end{table*}

In contrast, Hybrid Augmentation simulates a collaborative or meta-review setting. Here, the model receives the original human review along with four independently generated LLM reviews of the same manuscript. Rather than expanding a single critique, the model synthesizes multiple perspectives into a unified review. It is instructed to preserve the human reviewer’s core insights while integrating valid arguments introduced by the AI reviewers that were absent from the human review. This results in a genuinely hybrid idea provenance, reflecting contributions from both human and multiple AI reviewers. As in all transformation regimes, the final surface realization is fully AI-generated~\cite{hossain2025llms}.

\subsection{Dataset Statistics and Labeling Schema}

The final dataset comprises 20,690 usable reviews. Table~\ref{tab:dataset_stats} summarizes the distribution across generation types.
Each instance in the dataset is annotated with the following tuple:
\begin{equation}
D_i = \{T, O_{idea}, O_{text}, M, G, Meta\}
\end{equation}
where $T$ denotes the review text. $O_{idea}$ indicates the origin of evaluative reasoning and takes values {Human, AI, Mixed}.
$O_{text}$ denotes the origin of surface realization and takes values {Human, AI}. 
$M$ specifies the generating model (or \texttt{N/A} for human-authored text).
 $G$ denotes the generation regime (human, fully\_synthetic, rewritten, expanded, extract\_regenerate, hybrid) and  $Meta$ contains paper metadata, including venue, year, and decision outcome.
Note that $O_{text} = \text{Human}$ only for original source reviews.
\section{Experimental Setup}

\subsection{Benchmarking LLM Detection Methods}
\label{sec:benchmark}

To evaluate LLM detection behavior, we benchmark seven representative methods spanning four fundamentally different paradigms: likelihood-based, likelihood-ratio, perturbation-based, embedding-based, and supervised classification. Specifically, we include GLTR ~\cite{gehrmann2019gltr}, DetectGPT ~\cite{mitchell2023detectgpt}, Fast-DetectGPT ~\cite{bao2024fastdetectgpt}, Lastde++ ~\cite{xu2025trainingfree}, Binoculars ~\cite{hans2024binoculars}, RADAR ~\cite{hu2023radarrobustaitextdetection}, and the context-aware Anchor detector ~\cite{Yu2025IsYP}.

These methods differ categorically in how they characterize machine-generated text. Likelihood-based detectors (GLTR) analyze token rank distributions under a pretrained language model, assuming LLM outputs overproduce high-probability tokens. Likelihood-ratio methods (Binoculars) contrast scores from two related language models to reveal systematic generation discrepancies. Perturbation based approaches (DetectGPT, Fast-DetectGPT, Lastde++) measure changes in model likelihood under controlled perturbations, exploiting the observation that synthetic text often lies near local likelihood maxima. In contrast, supervised classifiers (RADAR) learn to discriminate human from AI text directly from labeled data. Finally, context-aware embedding detectors (Anchor) compare candidate reviews against manuscript-conditioned LLM-generated references using semantic similarity, incorporating document context beyond surface fluency.
All detectors are evaluated in their original pretrained or off-the-shelf configurations, without fine-tuning on \system. For score-based methods without predefined thresholds, we calibrate decision boundaries on a balanced held-out subset and fix them across experiments. For implementation specifics and algorithmic details, we refer readers to the original papers. Our repository includes full reproduction scripts and standardized implementations of all baselines.

\textbf{Ground Truth and Evaluation Protocol.}
For binary evaluation, we define two strict ground-truth classes: original human-written reviews (Human) and fully synthetic LLM-generated reviews (AI). Hybrid regimes are excluded from threshold calibration and standard accuracy computation and are instead used to assess robustness under mixed-provenance conditions.
For detectors that output probabilities, we follow the thresholds recommended in their original implementations. For score-based methods without prescribed thresholds (Anchor and Lastde++), we calibrate the decision boundary on a balanced held-out subset and fix it for all subsequent experiments. We report accuracy and confusion matrices separately for each provenance regime.

\subsection{Stylistic and Semantic Analysis}
\label{sec:stylistic}

Beyond binary detection accuracy, we characterize the systematic differences across provenance regimes (Human, Fully Synthetic, and Transformed) by analyzing their stylistic and semantic properties. These metrics illuminate how authorship origin impacts linguistic structure, rhetorical patterns, and semantic alignment.

\noindent\textbf{Lexical diversity and readability.}
We assess vocabulary richness using the Type-Token Ratio (TTR), computed directly from token counts without external libraries. Additionally, we quantify syntactic complexity and readability using the Flesch Reading Ease score, computed via the \texttt{textstat} toolkit.\footnote{\url{https://github.com/shivam5992/textstat}}

\noindent\textbf{Reviewer voice and interaction signals.}
Beyond surface fluency, to gauge rhetorical engagement, we analyze markers of reviewer agency and engagement. The use of first-person pronouns (e.g., “I”, “we”) often signals ownership, subjective judgment, and personal responsibility in evaluation. Reviews that explicitly adopt such voice may reflect stronger intellectual commitment or evaluative confidence. We also measure interrogative engagement by counting the number of questions posed in a review. Asking clarifying or critical questions can indicate deeper scrutiny and active reasoning about the manuscript. To quantify this, we use a pretrained sentence-level classifier\footnote{\url{https://huggingface.co/shahrukhx01/question-vs-statement-classifier}}
 to identify interrogative sentences and compute question frequency per review. These signals help characterize differences in rhetorical style and engagement between human-authored and AI-generated reviews.
 
\noindent\textbf{Citation and reference behavior.}
We quantify manuscript grounding through two complementary signals. First, we measure \textit{external citation count}, capturing references to prior work (e.g., bracketed citations or author-year mentions), which reflect how often the reviewer situates the paper within the broader literature. Second, we compute \textit{explicit manuscript reference count}, identifying direct pointers to the submitted paper such as “Section 3,” “Figure 2,” or “Table 1.” These markers indicate close engagement with specific parts of the manuscript. Both metrics are extracted using regex patterns tailored to common academic writing conventions.

\noindent\textbf{Semantic similarity and provenance alignment.}
To analyze semantic shifts across regimes, we embed all reviews with \texttt{gte\-\allowbreak multilingual\-\allowbreak base}
~\cite{zhang2024mgte} and compute pairwise cosine similarity.
We evaluate five comparison settings: (i) \textit{Human-Transformed}, measuring semantic preservation under stylistic modification; (ii) \textit{Human-LLM}, capturing divergence between human and independently generated machine reviews; (iii) \textit{Human-Human}, establishing a baseline for natural reviewer variation; (iv) \textit{LLM-Transformed (same model)}, assessing whether transformations move reviews toward a machine subspace; and (v) \textit{LLM-LLM (same model)}, measuring intra-model consistency.

\section{Results}
\begin{figure*}[!ht]
\centering

\includegraphics[width=0.9\textwidth]{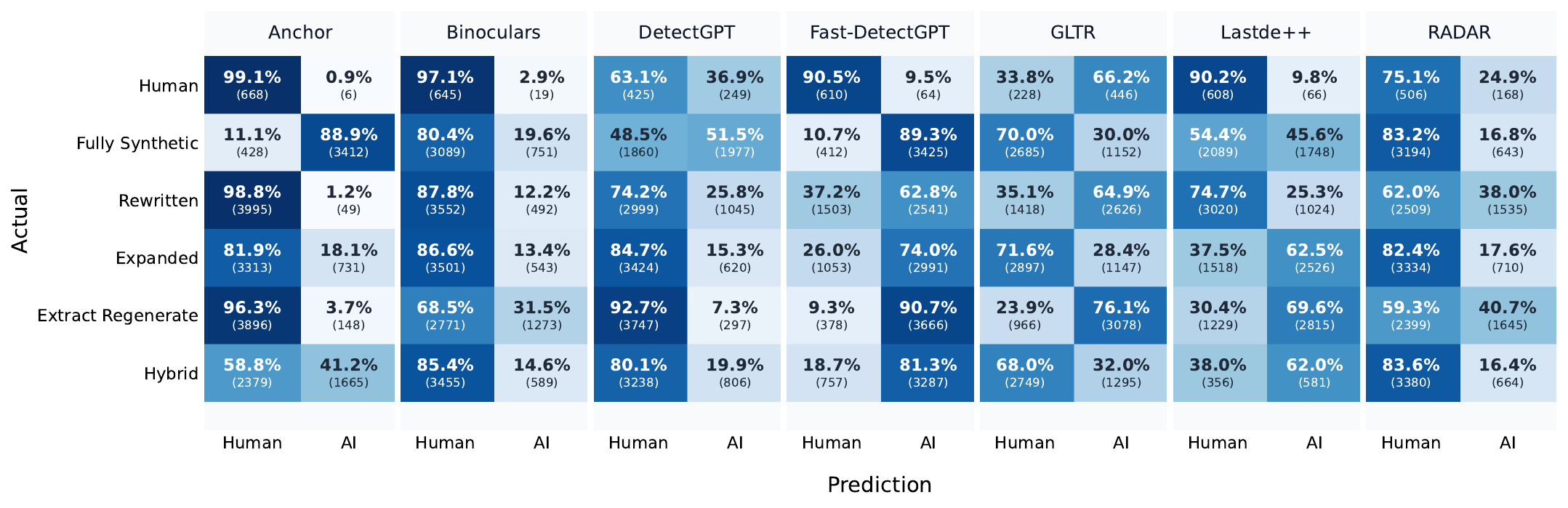}
\vspace{-1em}
\caption{Detector prediction breakdown by review generation regimes. Percentages are row-normalized per method. 
}

\label{fig:confusionMatrix}

\end{figure*}

\subsection{Results on Fully Synthetic Reviews}
\label{sec:results_grounded}

We begin with a sanity-check evaluation on the standard binary task of distinguishing original human-written reviews from fully synthetic LLM-generated reviews (Sec.~\ref{sec:fullgen}). Although existing detectors are primarily designed for general LLM text detection and are not optimized for the structured, domain-specific nature of peer review, this setting provides a clean baseline for grounding their behavior before moving to the more nuanced provenance-controlled regimes in Sec.~\ref{sec:mixed}.
Across detectors, performance varies. Likelihood-based methods such as {GLTR}, likelihood-ratio approaches such as {Binoculars}, and the supervised classifier {RADAR} degrade sharply on modern LLM-generated reviews as shown in Table~\ref{tab:accuracy_by_model}. Although these methods were previously validated on generic generation benchmarks, their reliance on token-level probability artifacts appears brittle in this structured, domain-specific setting \cite{tufts2025practical}.
In contrast, curvature-based detectors ({DetectGPT}, {Fast-DetectGPT}, {Lastde++}) and the embedding-based method ({Anchor}) demonstrate greater robustness. Instead of relying on surface-level likelihood statistics, curvature-based methods analyze the local geometry of the likelihood landscape under perturbations, while Anchor leverages manuscript-conditioned semantic similarity.

\paragraph{Impact of LLM on baselines.}
To further examine detector behavior, we disaggregate performance by source model (Table~\ref{tab:accuracy_by_model}). The results indicate that LLM detection reliability is strongly generator-dependent. Reviews produced by certain model families such as \texttt{GPT-5}, \texttt{Gemini-2.5}, and \texttt{Claude-Haiku-4.5} lead to larger performance drops in likelihood-based and supervised detectors compared to others. Rather than attributing this solely to model recency or scale, the results suggest that differences in fluency, stylistic calibration, and token distribution patterns can weaken signals relied upon by token-probability heuristics.
Robustness also varies across detectors. Curvature-based methods such as DetectGPT and Lastde++ remain effective overall but exhibit noticeable generator-specific variance. In contrast, Anchor and Fast-DetectGPT show more stable behavior across model families. Together, these findings highlight that LLM text detection is sensitive to generation characteristics and must be evaluated across diverse model regimes.

\subsection{Results on Provenance-Controlled Transformations}
\label{sec:results_transformed}

While the previous section established that certain detectors can reliably distinguish fully human from fully synthetic text, we now examine how these methods cope with the gray area of \textit{provenance-controlled transformations} (rewritten, expanded, extract and regenerated and hybrid reviews). This setting challenges the rigid binary assumption of prior work.
\begin{table*}[t]
\centering
\small
\setlength{\tabcolsep}{6pt}
\caption{Mean stylometric and discourse feature values across human and LLM-derived review types.}
\begin{tabular}{lcccccc}
\hline

& \multicolumn{2}{c}{\textbf{Language \& Style}}
& \multicolumn{2}{c}{\textbf{Voice \& Interaction}}
& \multicolumn{2}{c}{\textbf{Attribution \& Referencing}} \\

\textbf{Source}
& \textbf{Lexical Diversity}
& \textbf{Readability}
& \textbf{First-Person}
& \textbf{Questions}
& \textbf{Citations}
& \textbf{References} \\

\hline

Human & 0.55 & 37.84 & 5.04 & 2.34 & 0.95 & 1.57 \\

Fully Synthetic & 0.61 & 13.99 & 0.37 & 3.68 & 0.31 & 1.23 \\

\hline

Rewritten & 0.63 & 17.29 & 1.94 & 1.48 & 0.91 & 1.71 \\

Expanded & 0.51 & 26.50 & 3.52 & 2.33 & 2.23 & 5.96 \\

Extract Regenerate & 0.55 & 15.15 & 1.98 & 0.70 & 0.53 & 1.58 \\

Hybrid & 0.54 & 19.01 & 1.92 & 3.55 & 0.95 & 2.68 \\

\hline

\textbf{Transformed (mean)} & \textbf{0.56} & \textbf{19.49} & \textbf{2.34} & \textbf{2.02} & \textbf{1.16} & \textbf{2.98} \\

\hline
\end{tabular}

\label{tab:stylometric_features_mean}
\end{table*}




\begin{table}[t]
\centering
\small
\setlength{\tabcolsep}{4pt}

\caption{Average pairwise cosine similarity across  different settings using \texttt{gte-multilingual-base}.}
\vspace{-1em}
\label{tab:semantic_similarity_summary}

\begin{tabular}{lc}
\toprule
\textbf{Comparison} & \textbf{Similarity} \\
\midrule

\multicolumn{2}{c}{\textit{Reviewer consistency}} \\
Human $\leftrightarrow$ Human (same paper) & 0.83 \\
Fully Synthetic $\leftrightarrow$ Fully Synthetic (same paper) & 0.92 \\

\midrule
\multicolumn{2}{c}{\textit{Transformation preservation}} \\
Human $\leftrightarrow$ Transformed & 0.92 \\
Synthetic $\leftrightarrow$ Transformed & 0.88 \\

\midrule
\multicolumn{2}{c}{\textit{Manuscript alignment}} \\
Human $\leftrightarrow$ Manuscript & 0.82 \\
Fully Synthetic $\leftrightarrow$ Manuscript & 0.86 \\
Rewritten $\leftrightarrow$ Manuscript & 0.81 \\
Expanded $\leftrightarrow$ Manuscript & 0.81 \\
Extract Regenerate $\leftrightarrow$ Manuscript & 0.85 \\
Hybrid $\leftrightarrow$ Manuscript & 0.81 \\
\bottomrule
\end{tabular}
\end{table}

\paragraph{The breakdown of detector consensus.}
Figure~\ref{fig:confusionMatrix} reveals a sharp divergence once hybrid authorship is introduced. In the \textit{rewritten} regime, GLTR predicts 1,418 reviews (35.1\%) as Human and 2,626(64.9\%) as AI, while Fast-DetectGPT predicts 2,999 (74.2\%) as Human and 1,045 (25.8\%) as AI. 

The divergence intensifies in the \textit{expanded} regime. Fast-DetectGPT classifies 1,053 reviews (26.0\%) as Human and 2,991 (74.0\%) as AI, whereas Anchor predicts 3,313 (81.9\%) as Human and 731 (18.1\%) as AI. GLTR again differs, labeling 2,897 (71.6\%) as Human and 1,147 (28.4\%) as AI.

\paragraph{Implications for the binary paradigm.}
The core issue is not performance degradation; it is task fragmentation. Detectors that appear reliable under a strict Human vs.\ Fully-Synthetic binary split are in fact solving different implicit problems. Some detect surface realization (who wrote the words), while others detect semantic inheritance (who originated the evaluative reasoning).
As a result, “detection accuracy” on fully synthetic data is a poor predictor of real-world robustness. In realistic peer-review workflows—where text may be rewritten, expanded, or collaboratively augmented—the binary \textit{Human vs.\ AI} framing collapses. Authorship lies on a continuum of human-AI interaction that current binary detectors are not designed to model.

\subsection{Semantic Alignment}
\label{sec:results_semantic}

To decode the conflicting detector behaviors observed in Section~\ref{sec:results_transformed}, we turn to the underlying semantic architecture of the reviews. By analyzing embedding similarity (Table~\ref{tab:semantic_similarity_summary}), we uncover the structural reasons why different detection paradigms diverge.

\textit{The ``echo chamber'' of LLM generation.}
First, we observe a distinct difference in interpretative diversity. Human reviewers exhibit significant semantic variance (similarity score of 0.83), reflecting the natural diversity of human critique and focus. In contrast, LLM-generated reviews are far more homogenized, clustering tightly with a similarity of 0.92. When conditioned on the same manuscript, models tend to converge on similar points, lacking the idiosyncratic perspective of individual human experts.

\textit{Manuscript alignment.}
LLM-generated reviews exhibit higher manuscript similarity (0.86) than human reviews (0.82). Transformed reviews that do not access the manuscript directly retain similar similarity scores (0.81–0.85), indicating that manuscript-related semantic content is preserved through transformation. These results suggest that manuscript similarity is influenced both by direct conditioning on the manuscript and by semantic inheritance from the original human review.

\textit{Semantic inheritance.}
As shown in Table~\ref{tab:semantic_similarity_summary}, transformed reviews remain highly similar to their original human source (similarity 0.92) and are less similar to independently generated LLM reviews (similarity 0.88). This demonstrates semantic inheritance: LLM transformations preserve the semantic structure of the human-authored input.
In other words, even when a review is rewritten or expanded by an AI, it inherits the semantic structure, argumentative flow, and evaluative logic of the human author.


\subsection{Stylometric Characteristics Across Regimes}
\label{sec:results_stylometric}

Beyond semantic structure, we observe distinct ``fingerprints'' in the stylistic realization of reviews. Table~\ref{tab:stylometric_features_mean} quantifies these differences, revealing how machine generation systematically alters the rhetorical voice of critique.

\noindent\textbf{The erasure of subjectivity.}
The most profound shift occurs in authorial presence. Human reviews are characterized by frequent use of first-person pronouns (5.04 per review), reflecting a subjective, engaged evaluative stance. In contrast, LLM-generated reviews exhibit a marked "objectivity bias," reducing first-person usage to just 0.37. This creates a detached, formal tone that mimics an idealized, neutral arbiter rather than an engaged peer.

\noindent\textbf{Complexity and lexical ``smoothing.''}
This detachment is mirrored in vocabulary usage. Fully synthetic reviews display significantly higher lexical diversity (0.61 vs. 0.55), suggesting a probabilistic sampling that avoids the repetitive, focused vocabulary typical of human argumentation. Furthermore, readability scores indicate a shift toward complexity; LLMs produce text with a Flesch Reading Ease score of
 13.99, compared to 0.55 for humans, creating sophistication that may mask a lack of genuine critical depth.
\textbf{The ``Cyborg'' nature of transformed reviews.}
Transformed reviews occupy an intermediate, dimension-specific position in stylistic space. On features associated with rhetorical voice, such as first-person pronoun usage and readability, they shift substantially toward the machine profile. Pronoun usage decreases markedly, and readability moves closer to fully synthetic reviews, reflecting reduced authorial presence and increased formalization. In this sense, the subjective tone of the original reviewer becomes attenuated.

However, this stylistic shift does not occur uniformly across all dimensions. Citation and manuscript-referencing behavior remains strongly human-aligned. Transformed reviews preserve and often amplify explicit references to sections, figures, and tables, exceeding the original human reviews in manuscript pointers and far surpassing fully synthetic reviews.

This asymmetry reveals a hybrid signature: stylistic realization becomes machine-like, while structural engagement with the manuscript remains grounded in the human critique. Taken together, these results show that contemporary generation methods can overwrite the \textit{style} of a human reviewer while preserving the \textit{substantive grounding} of the original evaluative reasoning.
\vspace{-0.1em}
\section{Concluding Remarks}
In high-stakes domains such as peer review, it is critical to rigorously understand the structure of human-AI interaction. Our goal in this work is not to advocate for the use of LLMs in peer review, nor to normalize automated reviewing. Rather, we argue that as LLM assistance becomes increasingly present in scholarly workflows, the community must be equipped with principled methods to evaluate and characterize such collaboration.
As such, we introduce \system, a dataset of 20,690 reviews collected and generated from ICLR and NeurIPS, designed to disentangle idea provenance from text provenance. By benchmarking state-of-the-art LLM detection methods on \system, we show that LLM detection in peer review is fundamentally more complex than a binary Human-versus-AI classification problem. Existing detectors implicitly solve different tasks. While some capture statistical surface realization, others track semantic alignment. When evaluative ideas originate from humans but the surface text is AI-generated, detector predictions fragment and frequently contradict one another.
This fragmentation exposes a structural limitation of current attribution frameworks. Authorship in modern peer review cannot be reduced to a strict binary label. Instead, it lies on a continuum shaped by varying degrees of human reasoning and AI-mediated expression. Treating detection as a simple classification problem produces misleading notions of performance that fail under realistic hybrid conditions.

Moving forward, we argue that research should move beyond binary detection toward provenance quantification. Rather than asking “Human or AI?”, future systems should estimate degrees of semantic contribution, stylistic realization, and collaborative influence. Such multidimensional modeling does not endorse AI use in peer review; instead, it provides a more precise and responsible framework for evaluating its presence, understanding its impact, and preserving the integrity of scientific evaluation.



\bibliographystyle{ACM-Reference-Format}
\bibliography{sample-base}

\end{document}